\documentclass[english,letter]{article}
\usepackage[T1]{fontenc}
\usepackage[latin9]{inputenc}
\usepackage{color}
\usepackage{float}
\usepackage{bm}
\usepackage{amsmath}
\usepackage{graphicx}

\makeatletter

\providecommand{\tabularnewline}{\\}
\floatstyle{ruled}
\newfloat{algorithm}{tbp}{loa}
\providecommand{\algorithmname}{Algorithm}
\floatname{algorithm}{\protect\algorithmname}

\usepackage{aaai}
\usepackage{times}

\renewcommand\[{\begin{equation}}
\renewcommand\]{\end{equation}} 

\makeatother

\usepackage{babel}
\begin{document}

\title{Ordering-sensitive and Semantic-aware Topic Modeling}

\author{Min Yang \\ The University of Hong Kong \\ {\tt myang@cs.hku.hk} \And Tianyi Cui \\ Zhejiang University \\ {\tt tianyicui@gmail.com} \And Wenting Tu \\ The University of Hong Kong \\ {\tt wttu@cs.hku.hk}}
\maketitle
\begin{abstract}
\textcolor{black}{Topic modeling of textual corpora is an important
and challenging problem. In most previous work, the ``bag-of-words''
assumption is usually made which ignores the ordering of words. This
assumption simplifies the computation, but it unrealistically loses
the ordering information and the semantic of words in the context.
In this paper, we present a Gaussian Mixture Neural Topic Model (GMNTM)
which incorporates both the ordering of words and the semantic meaning
of sentences into topic modeling. Specifically, we represent each
topic as a cluster of multi-dimensional vectors and embed the corpus
into a collection of vectors generated by the Gaussian mixture model.
Each word is affected not only by its topic, but also by the embedding
vector of its surrounding words and the context. The Gaussian mixture
components and the topic of documents, sentences and words can be
learnt jointly. Extensive experiments show that our model can learn
better topics and more accurate word distributions for each topic.
Quantitatively, comparing to state-of-the-art topic modeling approaches,
GMNTM obtains significantly better performance in terms of perplexity,
retrieval accuracy and classification accuracy. }
\end{abstract}

\section{\textcolor{black}{Introduction }}

\textcolor{black}{With the growing of large collection of electronic
texts, much attention has been given to topic modeling of textual
corpora, designed to identify representations of the data and learn
thematic structure from large document collections without human supervision.
Topic models have been applied to a variety of applications, including
information retrieval \cite{wei2006lda}, collaborative filtering
\cite{marlin2003modeling}, authorship identification \cite{rosen2004author}
and opinion extraction \cite{lin2012weakly}, etc. Existing topic
models \cite{griffiths2004hierarchical,mcauliffe2008supervised,blei2012probabilistic}
are built based on the assumption that each document is represented
by a mixture of topics, where each topic defines a probability distribution
over words. These models, including the probabilistic latent semantic
analysis (PLSA) \cite{hofmann1999probabilistic} model and latent
Dirichlet allocation (LDA) \cite{blei2003latent} model, can be viewed
as graphical models with latent variables. Some non-parametric extensions
to these models have also been quite successful \cite{teh2006hierarchical,steyvers2007probabilistic}.
Nevertheless, exact inference for these model is computationally hard,
so one has to resort to slow or inaccurate approximations to compute
the posterior distribution over topics. New undirected graphical model
approaches, including the Replicated softmax model \cite{hinton2009replicated},
are also successfully applied to exploring the topics of the text,
and in particular cases they outperform LDA \cite{srivastava2013modeling}. }

\textcolor{black}{A major limitation of these topic models and many
of their extensions is the bag-of-word assumption, which assumes that
document can be fully characterized by bag-of-word features. This
assumption is favorable in the computational point of view, but loses
the ordering of the words and cannot properly capture the semantics
of the context. For example, the phrases ``the department chair couches
offers'' and ``the chair department offers couches'' have the same
unigram statistics, but are about quite different topics. When deciding
which topic generated the word ``chair'' in the first sentence,
knowing that it was immediately preceded by the word ``department''
makes it much more likely to have been generated by a topic that assigns
high probability to words related to university administration \cite{wallach2006topic}. }

\textcolor{black}{There has been little work on developing topic models
where the order of words is taken into consideration. To remove the
assumption that the order of words is negligible, \citeauthor{gruber2007hidden} \shortcite{gruber2007hidden}
propose modeling the topics of words in the document via a Markov
chain. \citeauthor{wallach2006topic} \shortcite{wallach2006topic}
explores a hierarchical generative probabilistic model that incorporates
both n-gram statistics and latent topic variables. Even though they
consider the order of words to some extent, their model is still not
capable of characterizing the semantics of words. For example, the
integer representation of the words ``teacher'' and ``teach''
are completely unrelated, even if we know they have strong semantic
connections and are very likely belonging to the same topic. To seek
a distributed way of representing words that capture semantic similarities,
several Neural Probabilistic Language Models (NPLMs) have been proposed
\cite{mnih2009scalable,mnih2012fast,mnih2013learning,mikolov2013efficient,le2014distributed}.
Nevertheless, the dense word embeddings learned by previous NPLMs
cannot be directly interpreted as topics. This is because that word
embeddings are usually considered opaque, in the sense that it is
difficult to assign meanings to the the vector representation.}

\textcolor{black}{In this paper, we proposed a novel topic model called
the Gaussian Mixture Neural Topic Model (GMNTM). The work is inspired
by the recent neural probabilistic language models \cite{mnih2009scalable,mnih2012fast,mnih2013learning,mikolov2013efficient,le2014distributed}.
We represent the topic model as a Gaussian mixture model of vectors
which encode words, sentences and documents. Each mixture component
is associated with a specific topic. We present a method that jointly
learns the topic model and the vector representation. As in NPLM methods,
the word embeddings are learnt to optimize the predictability of a
word using its surrounding words, with an important constraint that
the vector representations are sampled from the Gaussian mixture which
represents topics. Because the semantic meaning of sentences and documents
are incorporated to infer the topic of a specific word, in our model,
words with similar semantics are more likely to be clustered into
the same topic, and topics of sentences and documents are more accurately
learned. It potentially overcomes the weaknesses of the bag-of-word
method and the bag-of-n-grams method, both of which don't use the
order of words or the semantic of the context. We conduct experiments
to verify the effectiveness of the proposed model on two widely used
publicly available datasets. The experiment results show that our
model substantially outperforms the state-of-the-art models in terms
of perplexity, document retrieval quality and document classification
accuracy.}

\section{Related works}

In the past decade, a great variety of topic models have been proposed,
which can extract interesting topics in the form of multinomial distributions
automatically from texts \cite{blei2003latent,griffiths2004hierarchical,blei2012probabilistic,gruber2007hidden,hinton2009replicated}.
Among these approaches, LDA \cite{blei2003latent} and its variants
are the most popular models for topic modeling. The mixture of topics
per document in the LDA model is generated from a Dirichlet prior
mutual to all documents in the corpus. Different extensions of the
LDA model have been proposed. For example, \citeauthor{teh2006hierarchical} \shortcite{teh2006hierarchical}
assumes that the number of mixture components is unknown a prior and
is to be inferred from the data. \citeauthor{mcauliffe2008supervised} \shortcite{mcauliffe2008supervised}
develops a supervised latent Dirichlet allocation model (sLDA) for
document-response pairs. Recent work incorporates context information
into the topic modeling, such as time \cite{wang2006topics}, \textcolor{black}{geographic
location \cite{mei2006probabilistic},} authorship \cite{steyvers2004probabilistic},
and sentiment \cite{yang2014learning,yang2014topic}, to make topic
models fit expectations better.

Recently, there are several undirected graphical models being proposed,
which typically outperform LDA. \citeauthor{mcauliffe2008supervised} \shortcite{mcauliffe2008supervised}
present a two-layer undirected graphical model, called ``Replicated
Softmax'', that can be used to model and automatically extract low-dimensional
latent semantic representations from a large unstructured collection
of document. \citeauthor{hinton2009replicated} \shortcite{hinton2009replicated}
extend ``Replicated Softmax'' by adding another layer of hidden
units on top of the first with bipartite undirected connections. Neural
network based approaches, such as Neural Autoregressive Density Estimators
(DocNADE) \cite{larochelle2012neural} and Hybrid Neural Network-Latent
Topic Model \cite{wan2012hybrid}, are also shown outperforming the
LDA model. 

However, all of these these topic models employ the bag-of-words assumption,
which is rarely true in practice. The bag-of-word assumption loses
the ordering of the words and ignore the semantics of the context.\textcolor{green}{{}
}There are several previous literature taking the order of words into
account. \citeauthor{wallach2006topic} \shortcite{wallach2006topic}\textcolor{black}{{}
explores a hierarchical generative probabilistic model that incorporates
both n-gram statistics and latent topic variables. They extend a unigram
topic model so that it can reflect properties of a hierarchical Dirichlet
bigram model. }\citeauthor{gruber2007hidden} \shortcite{gruber2007hidden}\textcolor{black}{{}
propose modeling the topic of words a Markov chain. }\citeauthor{florezdeep} \shortcite{florezdeep}\textcolor{black}{{}
exploits the semantics regularities captured by a Recurrent Neural
Network (RNN) in text documents to build a recommender system. Although
these methods captures the ordering of words, none of them them consider
the semantics, thus they cannot capture the semantic similarities
between words such as ``teach'' and ``teacher''. In contrast,
our model is inspired by the recent work in learning vector representations
of words which are proved to capture the semantics of texts \cite{mnih2009scalable,mnih2012fast,mnih2013learning,mikolov2013efficient,le2014distributed}.
Our topic model captures both the ordering of words and the semantics
of the context. As a consequence, semantically similar words are more
likely having similar topic distribution (e.g., ``Jesus'' and ``Christ''
).}

\section{\textcolor{black}{The GMNTM Model}}

In this section, we first describe the GMNTM model as a probabilistic
generative model. Then we illustrate the inference algorithm for estimating
the model parameters.

\subsection{Generative model}

We assume there are $W$ different words in the vocabulary and there
are $D$ documents in corpus. For each word $w\in\{1,\ldots,W\}$
in vocabulary, there is an associated $V$-dimensional vector representation
${\rm vec}(w)\in{\cal R}^{V}$ for the word. Each document in corpus
with index $d\in\{1,\ldots,D\}$ also has a vector representation
${\rm vec}(d)\in{\cal R}^{V}$. If all the documents contain $S$
sentences, then these sentences are indexed by $s\in\{1,\ldots,S\}$.
The sentence with index $s$ is associated with a vector representation
${\rm vec}(s)\in{\cal R}^{V}$.

There are $T$ topics in the GMNTM model, where $T$ is designated
by the user. Each topic corresponds to a Gaussian mixture component.
The $k$-th topic is represented by a $V$-dimensional Gaussian distribution
${\cal N}(\mu_{k},\Sigma_{k})$ with mixture weight $\pi_{k}\in\mathcal{R}$,
where $\mu_{k}\in{\cal R}^{V}$, $\Sigma_{k}\in{\cal R}^{V\times V}$,
and $\sum_{k=1}^{T}\pi_{k}=1$. The parameters of the Gaussian mixture
model are collectively represented by
\[
\lambda=\{\pi_{k},\mu_{k},\Sigma_{k}\}\quad k=1,\ldots,T
\]
Given the collection of parameters, we use

\begin{equation}
p({\bf x}|\lambda)=\sum_{i=1}^{T}\pi_{i}{\cal N}({\bf x}|\mu_{i},\Sigma_{i})\label{eq:gmm-specification}
\end{equation}
to represent the probability distribution for sampling a vector ${\bf x}$
from the Gaussian mixture model.

We describe the procedure that the corpus is generated. Given the
Gaussian mixture model $\lambda$, the generative process  is described
as follow: for each word $w$ in the vocabulary, we sample its topic
$z(w)$ from the multinomial distribution $\pi:=(\pi_{1},\pi_{2},\dots,\pi_{T})$
and sample its vector representation ${\rm vec}(w)$ from distribution
${\cal N}(\mu_{z(w)},\Sigma_{z(w)})$. Equivalently, the vector ${\rm vec}(w)$
is sampled from the Gaussian mixture model parameterized by $\lambda$.
For each document $d$ and each sentence $s$ in the document, we
sample their topics $z(d)$, $z(s)$ from distribution $\pi$ and
sample their vector representations, namely ${\rm vec}(d)$ and ${\rm vec}(s)$,
also from the Gaussian mixture model. Let $\varPsi$ be the collection
of latent vectors associated with all the words, sentences and documents
in the corpus, 
\[
\varPsi:=\left\{ {\rm vec}(w)\right\} \cup\left\{ {\rm vec}(d)\right\} \cup\left\{ {\rm vec}(s)\right\} 
\]

For each word slot in the sentence, its word realization is generated
according to the document's vector ${\rm vec}(d)$, the current sentence's
vector ${\rm vec}(s)$ as well as at most $m$ previous words in the
same sentence. Formally, for the $i$-th location in the sentence,
we represent its word realization by $w_{i}$. The probability distribution
of $w_{i}$ is defined by:
\begin{align}
 & p\left(w_{i}=w|d,s,w_{i-m},\ldots,w_{i-1}\right)\nonumber \\
\propto & \exp(a_{{\rm doc}}^{w}+a_{{\rm sen}}^{w}+\sum_{t=1}^{m}a_{t}^{w}+b)\label{eq:p(w)}
\end{align}
where $a_{{\rm doc}}$, $a_{{\rm sen}}$ and $a_{t}$ are influences
from the document, the sentence and the previous word, respectively.
They are defined by 
\begin{align}
a_{{\rm doc}}^{w} & =\langle u_{{\rm doc}}^{w},{\rm vec}(d)\rangle\label{eq:alpha-1}\\
a_{{\rm sen}}^{w} & =\langle u_{{\rm sen}}^{w},{\rm vec}(s)\rangle\label{eq:alpha-2}\\
a_{t}^{w} & =\langle u_{t}^{w},{\rm vec}(w_{i-t})\rangle\label{eq:alpha-3}
\end{align}
Here, $u_{{\rm doc}}^{w},u_{{\rm sen}}^{w},u_{t}^{w}\in\mathcal{R}^{V}$
are parameters of the model, and they are shared across all slots
in the corpus. We use $U$ to represent this collection of vectors,
\[
U:=\{u_{{\rm doc}},u_{{\rm sen}}\}\cup\left\{ u_{t}|t\in1,2,\ldots,m\}\right\} 
\]

Combining the equations above, the probability distribution of $w_{i}$
is defined by a multi-class logistic model, where the features come
from the vectors associated with the document, the sentence and the
$m$ previous words. By estimating the model parameters, we learn
the word representations that make one word predictable from its previous
words and the context. Jointly, we learn the distribution of topics
that words, sentences and documents belong to.

Given the model parameters and the vectors for documents, sentences
and words, we can infer the posterior probability distribution of
topics. In particular, for a document $d$ with vector representation
${\rm vec}(d)$, the posterior distribution of its topic, namely $q(z(d))$,
is easy to calculate. For any $z\in{1,2,\dots,T}$, we have 
\[
q(z(d)=z)=\frac{\pi_{z}{\cal N}\left({\rm vec}(d)|\mu_{z},\Sigma_{z}\right)}{\sum_{k=1}^{T}\pi_{k}{\cal N}\left({\rm vec}(d)|\mu_{k},\Sigma_{k}\right)}.
\]
Similarly, for each sentence $s$ in the document $d$, the posterior
distribution of its topic is
\[
q(z(s)=z)=\frac{\pi_{z}{\cal N}\left({\rm vec}(s)|\mu_{z},\Sigma_{z}\right)}{\sum_{k=1}^{T}\pi_{k}{\cal N}\left({\rm vec}(s)|\mu_{k},\Sigma_{k}\right)}.
\]
For each word $w$ in the vocabulary, the posterior distribution of
its topic is similarly calculated as

\[
q(z(w)=z)=\frac{\pi_{z}{\cal N}\left({\rm vec}(w)|\mu_{z},\Sigma_{z}\right)}{\sum_{k=1}^{T}\pi_{k}{\cal N}\left({\rm vec}(w)|\mu_{k},\Sigma_{k}\right)}
\]

Finally, for each word slot in the document, we also want to explore
its topic. Since the topic of a particular location in the document
is affected by its word realization and the sentence/document it belongs
to, we define the probability of it belonging to topic $z$ proportional
to the product of $q(z(w)=z)$, $q(z(s)=z)$ and $q(z(d)=z)$, where
$w$, $s$, and $d$ are the word, the sentence and the document that
this word slot associates with.

\subsection{Estimating model parameters}

We estimate the model parameters $\lambda$, $U$ and $\varPsi$ by
maximizing the likelihood of the generative model. The parameter estimation
consists of two stages. In Stage I, we maximize the likelihood of
the model with respect to $\text{\ensuremath{\lambda}}$. Since $\lambda$
characterizes a Gaussian mixture model, this procedure can be implemented
by the Expectation Maximization (EM) algorithm. In Stage II, we maximize
the model likelihood with respect to $U$ and $\varPsi$, this procedure
can be implemented by stochastic gradient descent. We alternatively
execute Stage I and Stage II until the parameters converge. The algorithm
in this section is summarized in Algorithm 1.

\subsubsection{Stage I: Estimating $\lambda$}

In this stage, the latent vector of words, sentences and documents
are fixed. We estimate the parameters of the Gaussian mixture model
$\lambda=\{\pi_{k},\mu_{k},\Sigma_{k}\}$. This is a classical statistical
estimation problem which can be solved by running the EM algorithm.
The reader can refer to the book \cite{bishop2006pattern} for the
implementation details.

\subsubsection{Stage II: estimating $U$ and $\varPsi$}

When $\lambda$ is known and fixed, we estimate the model parameters
$U$ and the latent vectors $\varPsi$ by maximizing the log-likelihood
of the generative model. In particular, we iteratively sample a location
in the corpus, and consider the log-likelihood of the observed word
at this location. Let the word realization at location $i$ be represented
by $w_{i}$. The log-likelihood of this location is equal to
\begin{align}
J_{i}(U,\varPsi) & =\log(p(\varPsi|\lambda))+a_{{\rm doc}}^{w_{i}}+a_{{\rm sen}}^{w_{i}}+\sum_{t=1}^{m}a_{t}^{w_{i}}+b\nonumber \\
 & -\log\big(\sum_{w}\exp(a_{{\rm doc}}^{w}+a_{{\rm sen}}^{w}+\sum_{t=1}^{m}a_{t}^{w}+b)\big)
\end{align}
where $p(\varPsi|\lambda)$ is the prior distribution of parameter
$\varPsi$ in the Gaussian mixture model, defined by equation \eqref{eq:gmm-specification}.
The quantities $a_{{\rm doc}}^{w}$, $a_{{\rm sen}}^{w}$ and $a_{t}^{w}$
are defined in equations \eqref{eq:alpha-1}, \eqref{eq:alpha-2},
and \eqref{eq:alpha-3}. The objective function $J_{i}(U,\varPsi)$
involves all parameters in the collections $(U,\varPsi)$. Taking
the computation efficiency into consideration, we only update the
parameters associated with the word $w_{i}$. Concretely, we update
\begin{align}
{\rm vec}(w_{i}) & \leftarrow{\rm vec}(w_{i})+\alpha\frac{\partial J_{i}(U,\varPsi)}{\partial{\rm vec}(w_{i})}\\
u_{t}^{w_{i}} & \leftarrow u_{t}^{w_{i}}+\alpha\frac{\partial J_{i}(U,\varPsi)}{\partial u_{t}^{w_{i}}}
\end{align}
with $\alpha$ as the learning rate. Similarly, we update ${\rm vec}(s)$,
${\rm vec}(d)$ and $u_{{\rm doc}}^{w}$, $u_{{\rm sen}}^{w}$ using
the same gradient step, as they are parameters associated with the
current sentence and the current document. Once the gradient update
is accomplished, we sample another location to continue the update.
The procedure terminates when there are sufficient number of updates
performed, so that both $U$ and $\varPsi$ converge to fixed values.

\begin{algorithm}
\begin{itemize}
\item Inputs: A corpus containing $D$ documents, $S$ sentences, and a
vocabulary containing $W$ distinct words
\item Initialize parameters

\begin{itemize}
\item Randomly initialize the vectors $\varPsi$.
\item Initialize parameters $U$ with all-zero vectors.
\item Initialize Gaussian mixture model parameters with the standard normal
distribution ${\cal N}(\bm{0},{\rm diag}(1))$.
\end{itemize}
\item Repeat until converge

\begin{itemize}
\item Fixing parameters $U$ and $\varPsi$, run the EM algorithm to estimate
the Gaussian mixture model parameters $\lambda$.
\item Fixing the Gaussian mixture model $\lambda$, run stochastic gradient
descent to maximize the log-likelihood of the model with respect to
parameters $U$ and $\varPsi$.
\end{itemize}
\end{itemize}
\protect\caption{Inference Algorithm}
\end{algorithm}

\section{Experiments}

In this section, we evaluate our model on the 20 Newsgroups and the
Reuters Corpus Volume 1 (RCV1-v2) data sets. Followed the evaluation
in \cite{srivastava2013modeling}, we compare our GMNTM model with
the state-of-the-art topic models in perplexity, retrieval quality
and classification accuracy.

\subsection{Datasets description}

We adopt two widely used datasets, the 20 Newsgroups data and the
RCV1-v2 data, in our evaluations. Data preprocessing is performed
on both datasets. We first remove non-alphabet characters, numbers,
pronoun, punctuation and stop words from the text. Then, stemming
is applied so as to reduce the vocabulary size and settle the issue
of data spareness. The detailed properties of the datasets are described
as follow.

\noindent \textbf{20 Newsgroups dataset: }This dataset is a collection
of 18,845 newsgroup documents%
\footnote{Available at http://qwone.com/\textasciitilde{}jason/20Newsgroups%
}. The corpus is partitioned into 20 different newsgroups, each corresponding
to a separate topic. Following the preprocessing in \cite{hinton2009replicated}
and \cite{larochelle2012neural}, the dataset is partitioned chronologically
into 11,314 training documents and 7,531 testing documents. 

\noindent \textbf{Reuters Corpus Volume 1 (RCV1-v2):} This dataset
is an archive of 804,414 newswire stories produced by Reuters journalists
between August 20, 1996, and August 19, 1997 \cite{lewis2004rcv1}%
\footnote{Available at http://trec.nist.gov/data/reuters/reuters.html%
}. RCV1-v2 has been manually categorized into 103 topics, and the topic
classes form a tree which is typically of depth 3. As in \cite{hinton2009replicated}
and \cite{larochelle2012neural}, the data was randomly split into
794,414 training documents and 10,000 testing documents.

\subsection{Baseline methods}

In the experiments, the proposed topic modeling approach is compared
with several baseline methods, which we describe below:

\noindent \textbf{Latent Dirichlet Allocation (LDA):} In the LDA model
\cite{blei2003latent}, we used the online variational inference implementation
of the gensim toolkit %
\footnote{http://radimrehurek.com/gensim/models/remodel.html%
}. We used the recommended parameter setting $\alpha=1/T$.

\noindent \textbf{Hidden Topic Markov Models (HMM):} This model is
proposed by \cite{gruber2007hidden}, which models the topics of words
in the document as a Markov chain. The HMM model is run using the
publicly available code%
\footnote{http://code.google.com/p/Oppenheimer/downloads/list%
}. \textcolor{black}{We use default settings for all hyper parameters.}

\noindent \textbf{Over-Replicated Softmax (ORS):} This model is proposed
by \cite{srivastava2013modeling}. It is a two hidden layer DBM model,
which has been shown to obtain a state-of-the-art performance in terms
of classification and retrieval tasks compared with Replicated Softmax
model \cite{hinton2009replicated} and Cannonade model \cite{larochelle2012neural}.

\subsection{Implementation details}

In our GMNTM model, the learning rate $\alpha$ is set to $0.025$
and gradually reduced to $0.0001$. For each word, at most $m=6$
previous words in the same sentence is used as the context. For easy
comparison with other models, the word vector size is set to the same
as the number of topics $V=T=128$. Increasing the word vector size
further could improve the quality of the topics that are generated
by the GMNTM model.

Documents are split into sentences and words using the NLTK toolkit
\cite{bird2006nltk}%
\footnote{http://www.nltk.org/%
}. The Gaussian mixture model is inferred using the variational inference
algorithm in scikit-learn toolkit \cite{pedregosa2011scikit}%
\footnote{http://scikit-learn.org/%
}. To perform comparable experiments with restricted vocabulary, words
outside of the vocabulary is replaced as a special token and doesn't
count into the word perplexity calculation.

\subsection{Generative model evaluation}

We first evaluate our model's performance as a generative model for
documents. We perform our evaluation on the 20 Newsgroups dataset
and the RCV1-v2 dataset. For each of the datasets, we extract the
words from raw data and preserve the ordering of words. We follow
the same evaluation as in \cite{srivastava2013modeling}, comparing
our model with the other models in terms of perplexity.

We estimate the log-probability fo\textcolor{black}{r 1000 held-out
documents tha}t are randomly sampled from the test sets. After running
the algorithm to infer the vector representations of words, sentences,
and documents in held-out test documents, the average test perplexity
per word is then estimated as $\exp\left(-\frac{1}{N}\sum_{w}\log p(w)\right)$,
where $N$ are the total number of words in the held-out test documents,
and $p(w)$ is calculated according to equation \eqref{eq:p(w)}.

Table 1 shows the perplexity for each dataset. The perplexity for
Over-Replicated Softmax is taken from \cite{srivastava2013modeling}.
As shown by Table 1, our model performs significantly better than
the other models on both datasets in terms of perplexity. More specifically,
for 20 Newsgroups data set, the perplexity decreases from 949 to 933,
and for RCV1-v2 data set, it decreases from 982 to 826. This verifies
the effectiveness of the proposed topic modeling approach in fitting
the dataset. The GMNTM model works particularly well on large-scale
datasets such as RCV1-v2.

\begin{table}
\renewcommand{\arraystretch}{1.5}

\begin{centering}
\begin{tabular}{|c|c|c|c|c|}
\hline 
Data Set & LDA & HTMM & ORS & GMNTM\tabularnewline
\hline 
20 Newsgroups & 1068 & 1013 & 949 & \textbf{933}\tabularnewline
\hline 
RCV1-v2 & 1246 & 1039 & 982 & \textbf{826}\tabularnewline
\hline 
\end{tabular}
\par\end{centering}

\protect\caption{Comparison of test perplexity per word with 128 topics}
\end{table}

\subsection{Document retrieval evaluation}

\begin{figure*}[!t]
\begin{centering}
\includegraphics[width=0.5\textwidth]{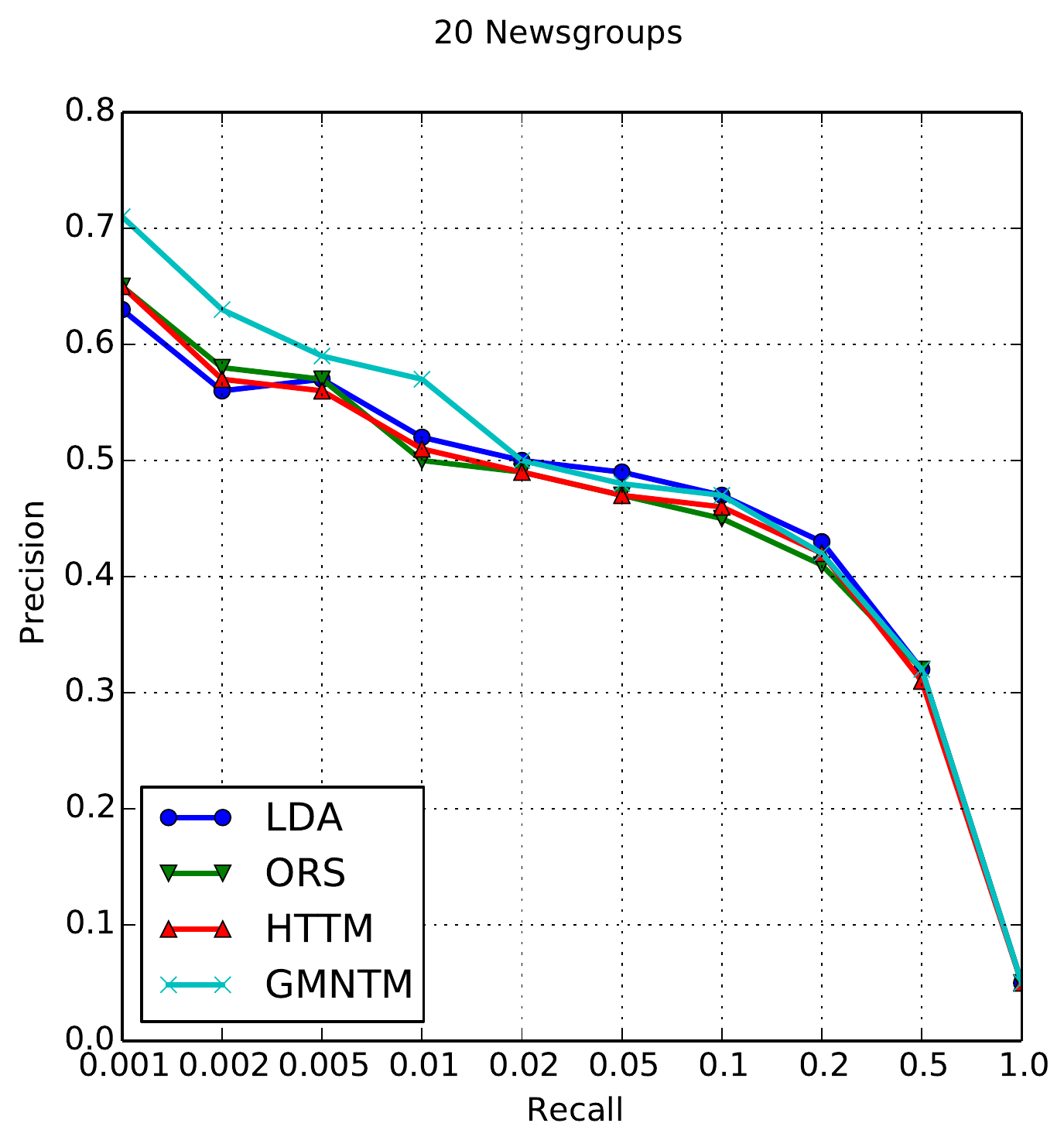}\includegraphics[width=0.5\textwidth]{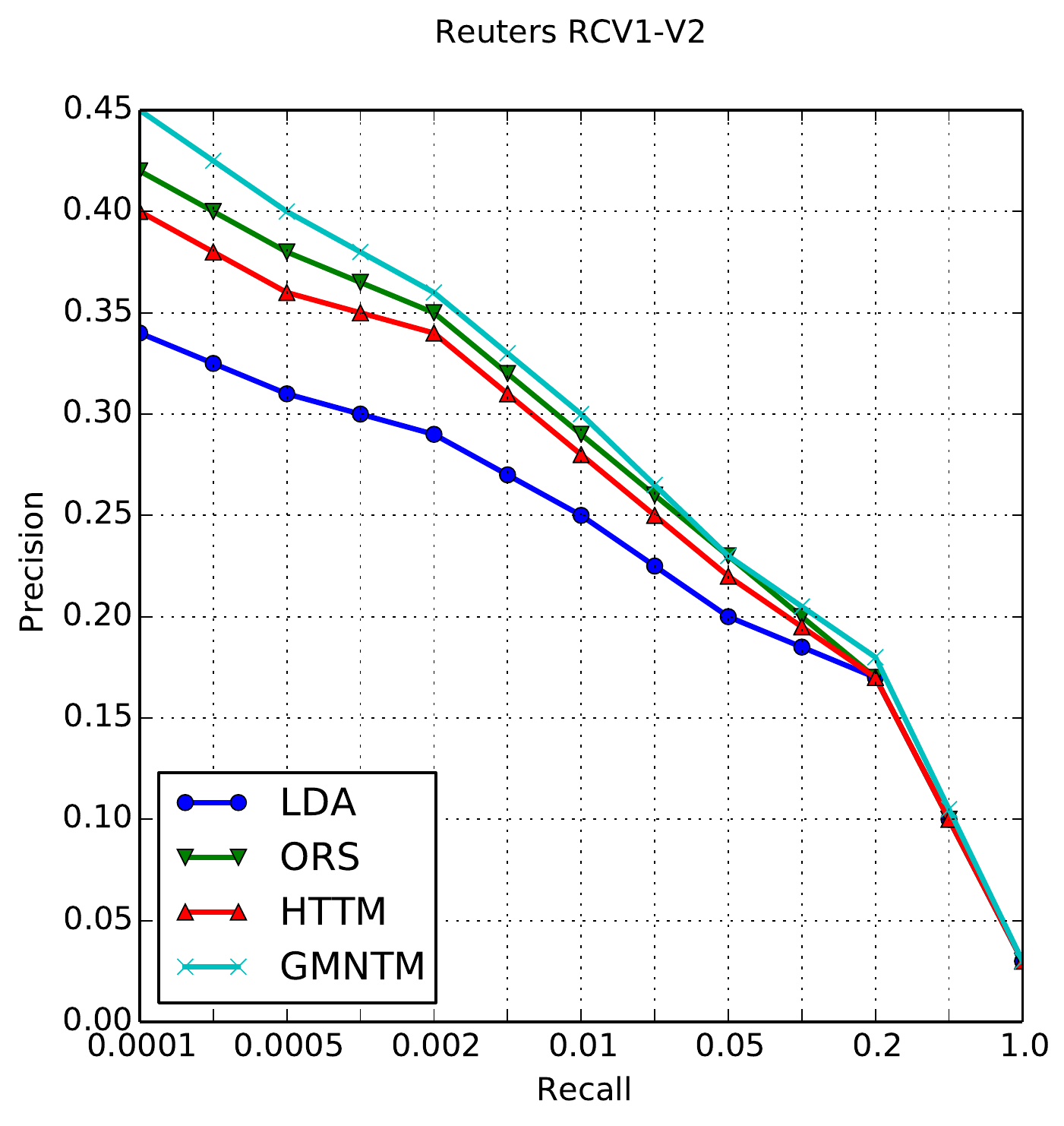}
\par\end{centering}

\protect\caption{Document Retrieval Evaluation}
\end{figure*}

To evaluate the quality of the documents representations that are
learnt by our model, we perform an information retrieval task. Following
the setting in \cite{srivastava2013modeling}, documents in the training
set are used as a database, while the test set is used as queries.
For each query, documents in the database are ranked using cosine
distance as the similarity metric. The retrieval task is performed
separately for each label and the results are averaged. Figure 1 compares
the precision-recall curves with 128 topics. The curves for LDA and
Over-Replicated are taken from \cite{srivastava2013modeling}.\textcolor{black}{{}
We see that for the 20 Newsgroups dataset, our model performs on par
or slightly better than the other models. While for the RCV1-v2 dataset,
our model achieves a significant improvement. Since RCV1-v2 contains
a greater amount of texts, the GMNTM model considering the ordering
of words is more powerful in mining the semantics of the text.}

\subsection{Document classification evaluation}

\begin{table}
\renewcommand{\arraystretch}{1.5}

\begin{centering}
\begin{tabular}{|c|c|c|c|c|}
\hline 
Data Set & LDA & HTMM & ORS & GMNTM\tabularnewline
\hline 
20 Newsgroups & 65.7\% & 66.5\% & 66.8\% & \textbf{73.1\%}\tabularnewline
\hline 
RCV1-v2 & 0.304 & 0.395 & 0.401 & \textbf{0.445}\tabularnewline
\hline 
\end{tabular}
\par\end{centering}

\protect\caption{Comparison of classification accuracy on 20 Newsgroups and Mean Precision
on Reuters RCV1-v2 with 128 topics}
\end{table}

Following the evaluation of \cite{srivastava2013modeling}, we also
perform document classification with the learnt topic representation
from our model. The same as in \cite{srivastava2013modeling}, multinomial
logistic regression with a cross entropy loss function is used for
the 20 Newsgroups data set, and the evaluation metric is the classification
accuracy. For the RCV1-v2 data set, we use independent logistic regression
for each label. The evaluation metric is Mean Average Precision.

We summarize the experiment results with 128 topics in Table 3. The
results of document classification for LDA and Over-Replicated Softmax
are taken from \cite{srivastava2013modeling}. According to Table
3, the proposed model substantially outperforms other models on both
datasets for document classification. For the 20 Newsgroups dataset,
the overall accuracy of the Over-Replicated Softmax model is 66.8\%,
which is slightly higher than LDA and HTMM. Our model further improves
the classification result to 73.1\%. On RCV1-v2 dataset, we observe
the similar results. The mean average precision increases from 0.401
(Over-Replicated Softmax) to 0.445 (our model).

\subsection{Qualitative inspection of topic specialization}

\begin{table*}[!t]
\renewcommand{\arraystretch}{1.3}

\begin{centering}
\begin{tabular}{|c|c|c|c|c|c|c|c|}
\hline 
\multicolumn{4}{|c|}{GMNTM Topic words} & \multicolumn{4}{c|}{LDA Topic Words}\tabularnewline
\hline 
god & space & game & key & god & space & year & key\tabularnewline
jesus & orbit & play & public & believe & nasa & hockey & encryption\tabularnewline
christ & earth & season & encryption & jesus & research & team & use\tabularnewline
believe & solar & team & security & sin & center & division & des\tabularnewline
christian & spacecraft & win & escrow & one & shuttle & league & system\tabularnewline
bible & surface & hockey & secure & mary & launch & nhl & rsa\tabularnewline
lord & planet & hand & data & lord & station & last & public\tabularnewline
truth & mission & series & privacy & would & orbit & think & security\tabularnewline
sin & satellite & chance & government & christian & april & maria & nsa\tabularnewline
faith & shuttle & nhl & nsa & accept & satellite & see & secure\tabularnewline
\hline 
\end{tabular}
\par\end{centering}

\protect\caption{Topic words}
\end{table*}

Since topic models are often used for the exploratory analysis of
unlabeled text, we also evaluate whether meaningful semantics are
captured by our model. Due to the space limit, we only illustrate
four topics extracted by our model and LDA which are topics about
religion, space, sports and security. These topics are also captured
as (sub)categories in the 20 Newsgroups dataset. Table 3 shows the
4 topics learnt by the GMNTM model and the corresponding topics learnt
by LDA. In each topic, we visualize it using 10 words with the largest
weights. The 4 topics shown in Table 3 for both models are easy for
interpretation according to the top words. However, we see that the
topics found by the two models are different in nature. GMNTM finds
topics that consist of the words that are consecutive in the document
or the words having similar semantics. For example, in the GMNTM model,
\textquotedblleft Christ\textquotedblright{} and \textquotedblleft christian\textquotedblright{}
share the same topics, mainly because they have strong semantic connections,
even though they don't co-occur that often, which makes LDA unable
to put them in the same topic. On the other hand, LDA often find some
general words such as \textquotedblleft would\textquotedblright{}
and \textquotedblleft accept\textquotedblright{} for the religion
topic, which are unhelpful for interpreting the topics.

\section{Conclusion and Future Work}

Rather than ignoring the semantics of the words and assuming that
the topic distribution within a document is conditionally independent,
in this paper, w\textcolor{black}{e introduce an ordering-sensitive
and semantic-aware topic modeling approach. The GMNTM model jointly
learns the topic of documents and the vectorized representation of
words, sentences and documents. The model learns better topics and
disambiguates words that belong to different topics. Comparing to
state-of-the-art topic modeling approaches, the GMNTM outperforms
in terms of perplexity, retrieval accuracy and classification accuracy.}

In future works, we will explore using non-parametric models to cluster
word vectors. For example, we look forward to incoporating infinite
Dirichelet process to automatically detect the number of topics. We
can also use hierarchical model to further capture the subtle semantics
of the text. As another promising direction, we consider building
topic models on popular neural probabilistic methods, such as the
Recurrent Neural Network Language Model (RNNLM). The GMNTM model has
appplications to several tasks in natural language processing, including
entity recognition, information extraction and sentiment analysis.
These applications also deserve further study,

\bibliographystyle{aaai}
\bibliography{topic_model}

\end{document}